\documentclass{article}
\PassOptionsToPackage{numbers, compress}{natbib}

\usepackage[final]{nips_2018}
\usepackage[hyphens]{url}
\usepackage[colorlinks]{hyperref}
\usepackage{wrapfig,booktabs}
\usepackage{url}
\usepackage{xcolor}
\usepackage{amsmath,amssymb,amsthm}
\usepackage[ruled]{algorithm2e}
\usepackage{wrapfig}
\usepackage{subcaption}
\usepackage{graphicx}
\usepackage{float}
\usepackage{mathtools, nccmath}
\hypersetup{breaklinks=true}

\title{Hallucinating Point Cloud into 3D Sculptural Object}
 \author{Chun-Liang Li$^{\star}$, Eunsu Kang$^{\star}$, Songwei Ge$^{\star}$, Lingyao Zhang, Austin Dill, \\
 {\bf Manzil Zaheer, Barnabas Poczos}\\
  Carnegie Mellon University\\\texttt{\small \{chunlial, eunsuk, songweig, lingyaoz, abdill, manzil, bapoczos\}@andrew.cmu.edu}\\
  $\star${\small Equal Contribution}
}

\begin{document}
\maketitle
\textit{}\vspace{-2.5em}
\section{Introduction}
Many artworks generated with the assistance of machine learning algorithms have appeared in the last three years.
Aesthetically impressive images have resulted that belong to the traditional art area of painting and have
received global recognition. The generated art images from~\citet{elgammal2017can} 
were favored by human viewers over the images of paintings at the prestigious Miami Art Basel. CloudPainter,
which was the top prize winner of the International Robot Art Competition in 2018, proved that a machine can achieve the
aesthetic level of professional painters~\citep{robotart2018}. 
Recently a painting that was generated by
GANs~\citep{goodfellow2014generative} 
appeared at the art auction by Christie’s~\citep{one_human}.
At the same time, there has been very little exploration in the area of 3D objects,
which traditionally would belong to the area of Sculpture. The creative generation of 3D Object research by Lehman et al. successfully
generated ``evolved'' forms, however, the final form was not far from the original form and it remained in the range of
mimicking the original~\citep{lehman2016creative}. Another relevant work, Neural 3D Mesh
Renderer~\citep{kato2018neural} focuses on adjusting the rendered mesh based on DeepDream~\citep{deepdream} textures. 

Our team of artists and machine learning researchers designed a creative algorithm that can generate authentic
sculptural artworks. These artworks do not mimic any given forms and cannot be easily categorized into the dataset
categories. Our approach extends DeepDream~\citep{deepdream} from images to 3D point clouds.  The proposed
algorithm, Amalgamated DeepDream (ADD), leverages the properties of point clouds to create objects with better
quality than the naive extension.  ADD presents promise for the creativity of machines, the kind of creativity that pushes
artists to explore novel methods or materials and to create new genres instead of creating variations of existing forms
or styles within one genre. For example, from Realism to Abstract Expressionism, or to Minimalism. 
Lastly, we present the sculptures that are 3D printed based on the point clouds created by ADD.


  \section{Learning to Generate Creative Point Clouds}
  Using deep learning to generate new objects has been studied in different data types, such as
  music~\citep{van2016wavenet}, images~\citep{goodfellow2014generative}, 3D
  voxels~\citep{wu2016learning} and point clouds~\citep{achlioptas2017learning, li2018point}.  
  However, these models are
  learned to generate examples from the ``same'' distribution of the given training data, instead of learning to
  generate ``creative'' objects~\citep{elgammal2017can}.  
  One alternative is to decode the combination
  \begin{wrapfigure}[7]{r}{0.34\textwidth}
  \vspace{-1.4em}
  \begin{subfigure}[b]{.485\linewidth}
  \includegraphics[width=\linewidth]{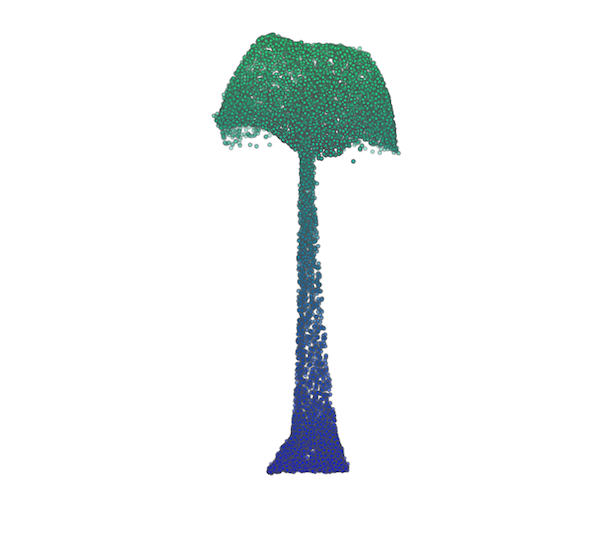}
  \vspace{-2.3em}
  \caption{Sampling}
  \end{subfigure}
  \hfill
  \begin{subfigure}[b]{.485\linewidth}
  \includegraphics[width=\linewidth]{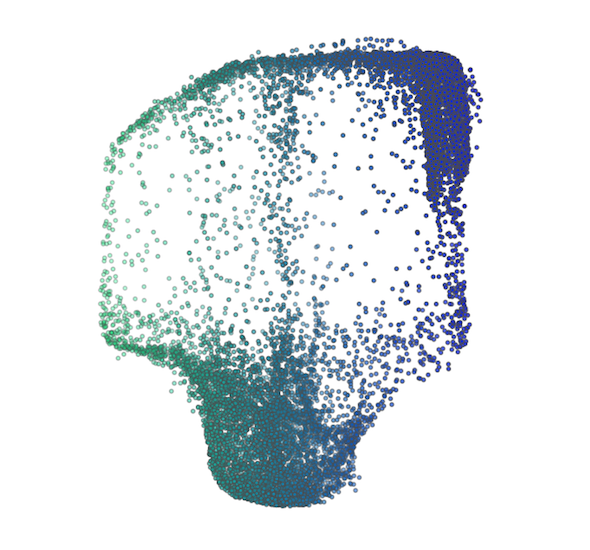}
  \vspace{-2.3em}
  \caption{Interpolation}
  \end{subfigure}
  \vspace{-1.5em}
  \caption{Results of~\citet{li2018point}.}
  \label{fig:pcgan}
  \end{wrapfigure} of latent codes of two objects in an autoencoder.  
  Empirical evidence shows that decoding mixed codes usually produces
  semantically meaningful objects with features from the corresponding objects. This approach has also been applied to image
  creation~\cite{carter2017using}. One could adopt 
  encoding-decoding algorithms for point clouds~\citep{yang2018foldingnet, groueix2018atlasnet, li2018point}
  to the same idea. The sampling and interpolation results based on~\citet{li2018point} are shown in Figure~\ref{fig:pcgan}.

  \paragraph{DeepDream for Point Clouds}
  In addition to simple generative models and interpolations, DeepDream leverages trained deep neural networks by enhancing
  patterns to create psychedelic and surreal images.  Given a trained neural network $f_\theta$ and an input image $x$, 
  DeepDream aims to modify $x$ to maximize (amplify) $f_\theta(x; a)$, where $a$ is an activation function of $f_\theta$.
  Algorithmically, 
  we iteratively modify $x$ via a gradient update  
  \useshortskip
  \begin{equation}
      x_t = x_{t-1} + \gamma \nabla_x f_\theta(x; a). 
      \label{eq:gradient}
  \end{equation}
  \begin{wrapfigure}[7]{r}{0.18\textwidth}
  \centering
  \vspace{-1.5em}
  \includegraphics[width=0.8\linewidth]{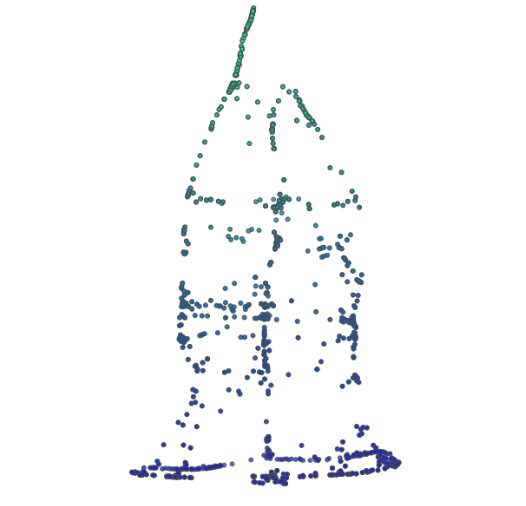}
  \caption{Naive DeepDream.}
  \label{fig:dd}
  \end{wrapfigure}
  One special case of DeepDream is when we use a classification network as $f_\theta$ and we replace $a$ with the outputs of the
  final layer corresponding to certain labels.  This is related to adversarial
  attack~\citep{szegedy2013intriguing} and unrestricted adversarial examples~\citep{unrestricted_Advex_2018,
  liu2018adversarial}. However, directly extending DeepDream to point clouds with neural networks for
  sets~\citep{qi2017pointnet, zaheer2017deep} results in undesirable point clouds with sparse areas as shown in
  Figure~\ref{fig:dd}. 

  \subsection{Amalgamated DeepDream (ADD)}
  To avoid sparsity in the generated point cloud, one compromise is to run DeepDream with fewer iterations or a smaller
  learning rate, but it results in limited difference from the input point cloud. 
  Another approach is conducting post-processing to add points to sparse areas. However, without prior knowledge of the
  geometry of the objects, making smooth post-processed results is non-trivial~\citep{hoppe1992surface}. 
  \begin{wrapfigure}[9]{r}{0.49\textwidth}
      \vspace{-5pt}
      \begin{algorithm}[H]
          \caption{Amalgamated DeepDream (ADD)}
          \SetKwInOut{Input}{input}\SetKwInOut{Output}{output}
          \Input{trained $f_\theta$ and input $x$, where $x_0=x$}
          \For{$t=1\dots T$} {
              $\hat{x} = x_{t-1} + \gamma \nabla_x f_\theta(x_{t-1}; a)$\\
              $x_t = \hat{x} \cup x$
          }
          \label{algo:add}
      \end{algorithm}
  \end{wrapfigure}

  \vspace{-8pt}
  As opposed to images, mixing two point clouds is a surprisingly simple task, which can be done by taking the union of two
  point clouds because of their permutation invariance property.
  We propose a novel DeepDream variant for point clouds with a set-union operation in the creation process
  to address the drawbacks of the naive extension. 
  To be specific, in each iteration after running the gradient update~\eqref{eq:gradient}, we
  \emph{amalgamate} the transformed point clouds with the input point clouds\footnote{We down-sample the point clouds
  periodically to avoid exploded number of points.}. We call the proposed algorithm Amalgamated
  DeepDream (ADD) as shown in Algorithm~\ref{algo:add}.
  A running example of ADD targeting the transformation of bottle into cone in ModelNet40~\citep{wu20153d}, which is the same as
  Figure~\ref{fig:dd}, is shown in Figure~\ref{fig:add}. 
  In each iteration, ADD enjoys the advantage of deforming objects based on gradient updates with respect to a trained neural network
  as DeepDream without creating obviously sparse areas. 
  In the meantime, it can better preserve the features of input point clouds
  with the amalgamation operation when we create new features based on DeepDream updates.
  More created point clouds are shown in Figure~\ref{fig:single}.

  \begin{minipage}{\linewidth}
      \vspace{-1.4em}
      \centering
      \begin{minipage}{0.32\linewidth}
          \begin{figure}[H]
              \begin{subfigure}[b]{.32\linewidth}
              \centering
              \includegraphics[trim={200 200 200 150},clip,width=\linewidth]{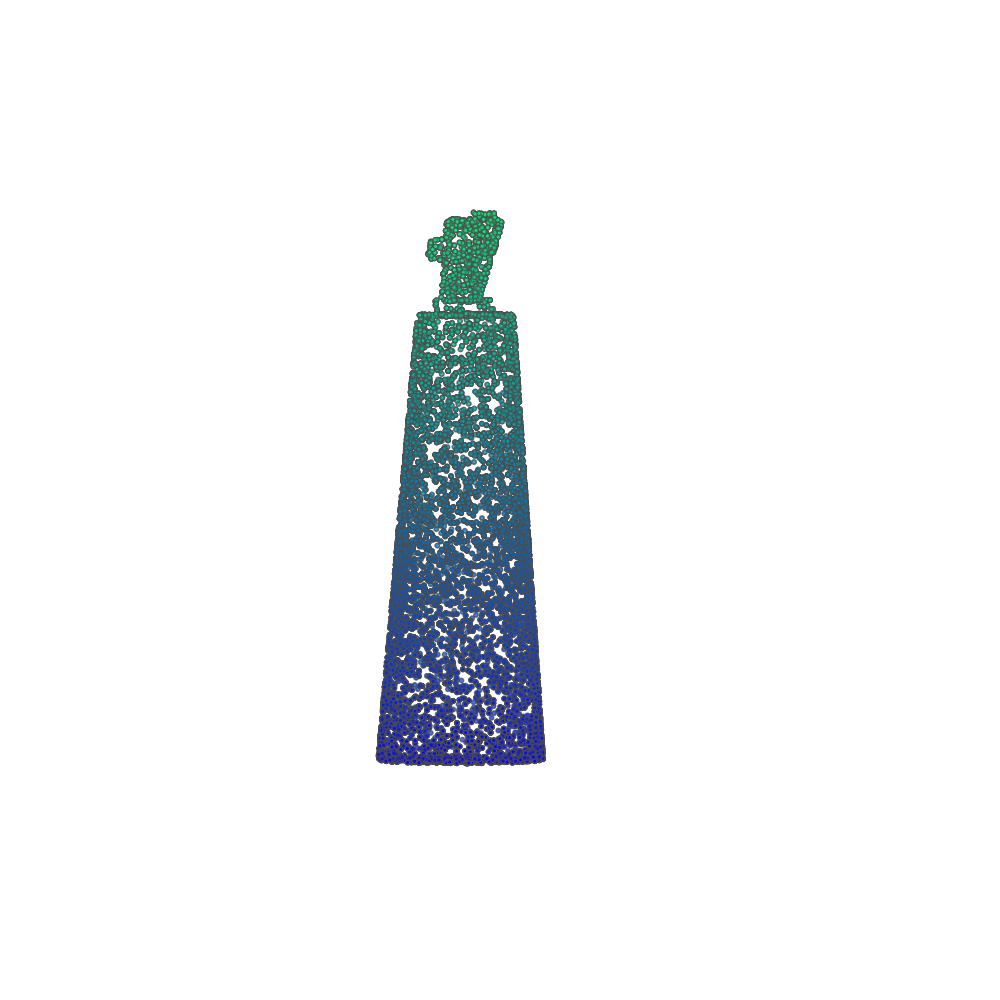}
              \caption{Input}
              \end{subfigure}
              \begin{subfigure}[b]{.32\linewidth}
              \centering
              \includegraphics[trim={200 200 200 150},clip,width=\linewidth]{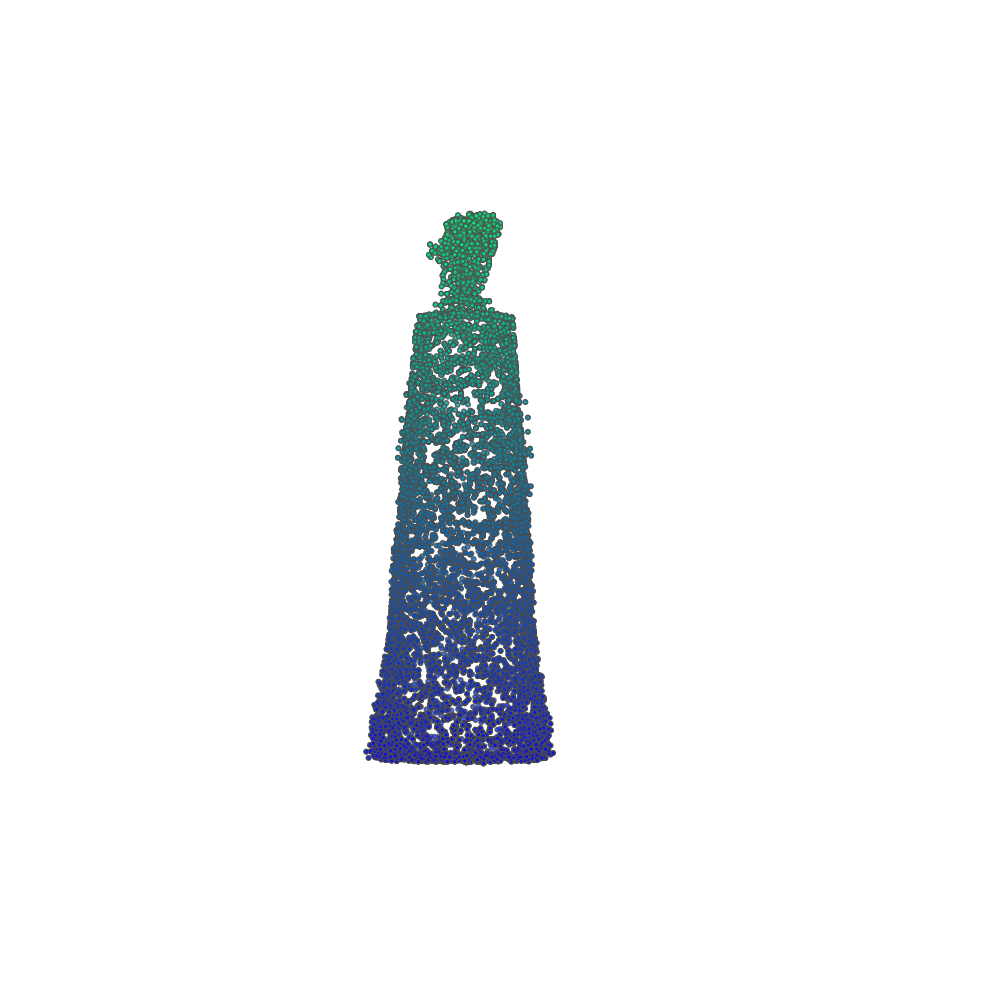}
              \caption{Iter. 5}
              \end{subfigure}
              \begin{subfigure}[b]{.32\linewidth}
              \centering
              \includegraphics[trim={200 200 200 150},clip,width=\linewidth]{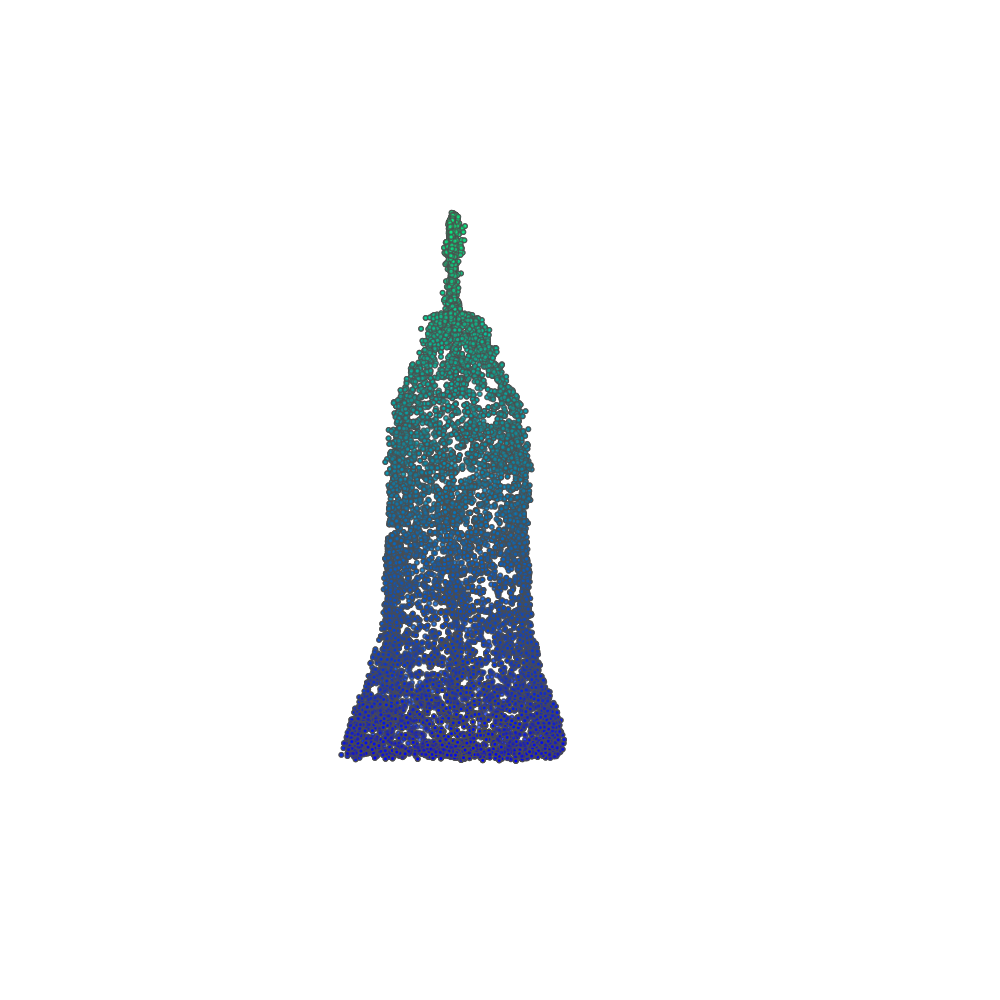}
              \caption{Iter. 10}
              \end{subfigure}
              \caption{Transforming a bottle into a cone via ADD.}
              \label{fig:add}
          \end{figure}
      \end{minipage}
      \hfill
      \begin{minipage}{0.32\linewidth}
          \begin{figure}[H]
              \centering
              \includegraphics[trim={200 200 200 150},clip,width=0.45\linewidth]{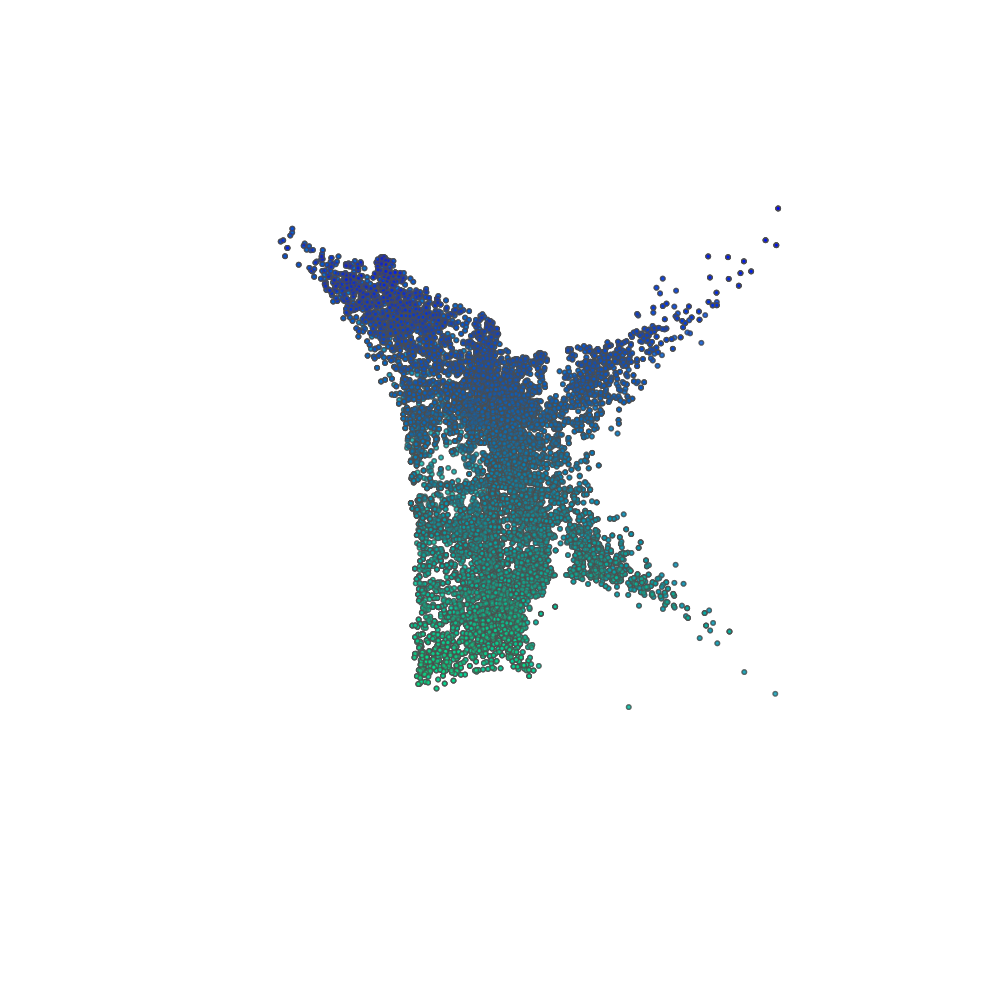}
              \hfill
              \includegraphics[trim={200 200 200 150},clip,width=0.45\linewidth]{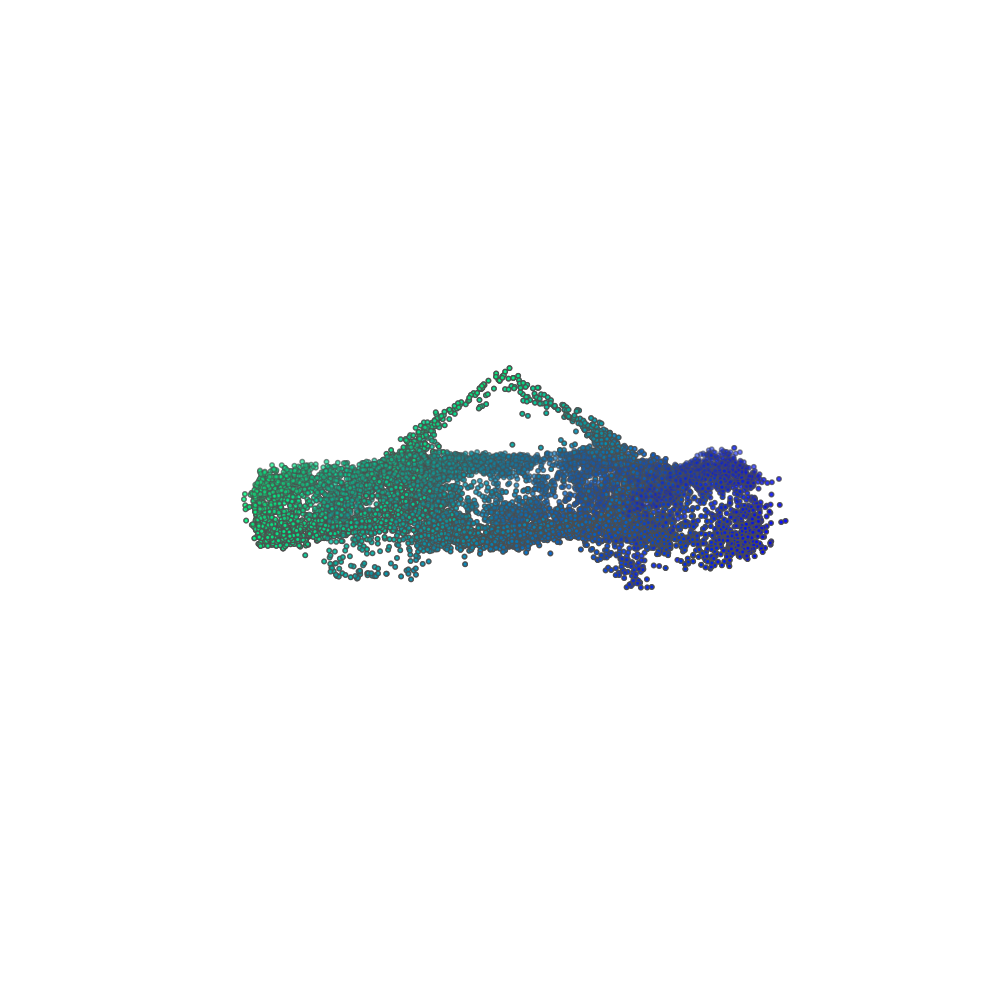}
              \caption{ADD with single object input.}
              \label{fig:single}
          \end{figure}
      \end{minipage}
      \hfill
      \begin{minipage}{0.32\linewidth}
          \begin{figure}[H]
              \centering
              \includegraphics[trim={200 200 200 150},clip,width=0.45\linewidth]{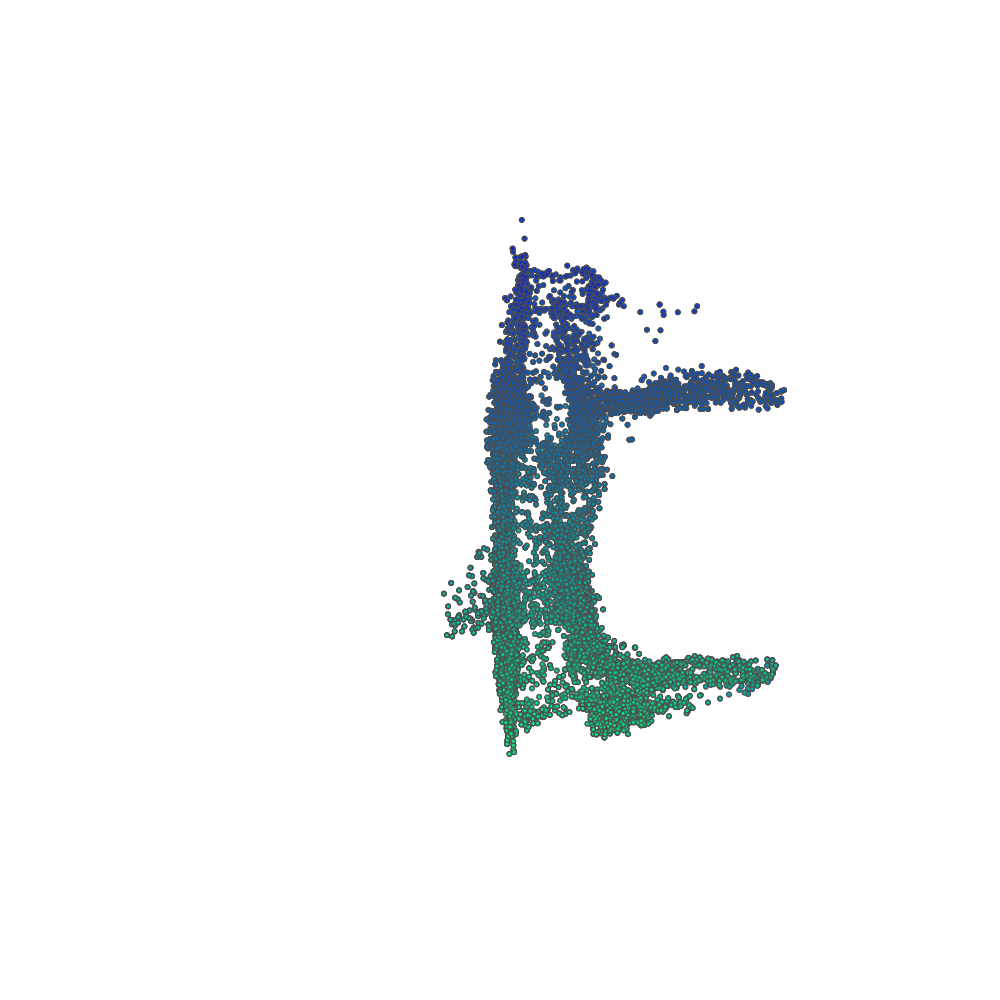}
              \hfill
              \includegraphics[trim={200 200 200 150},clip,width=0.45\linewidth]{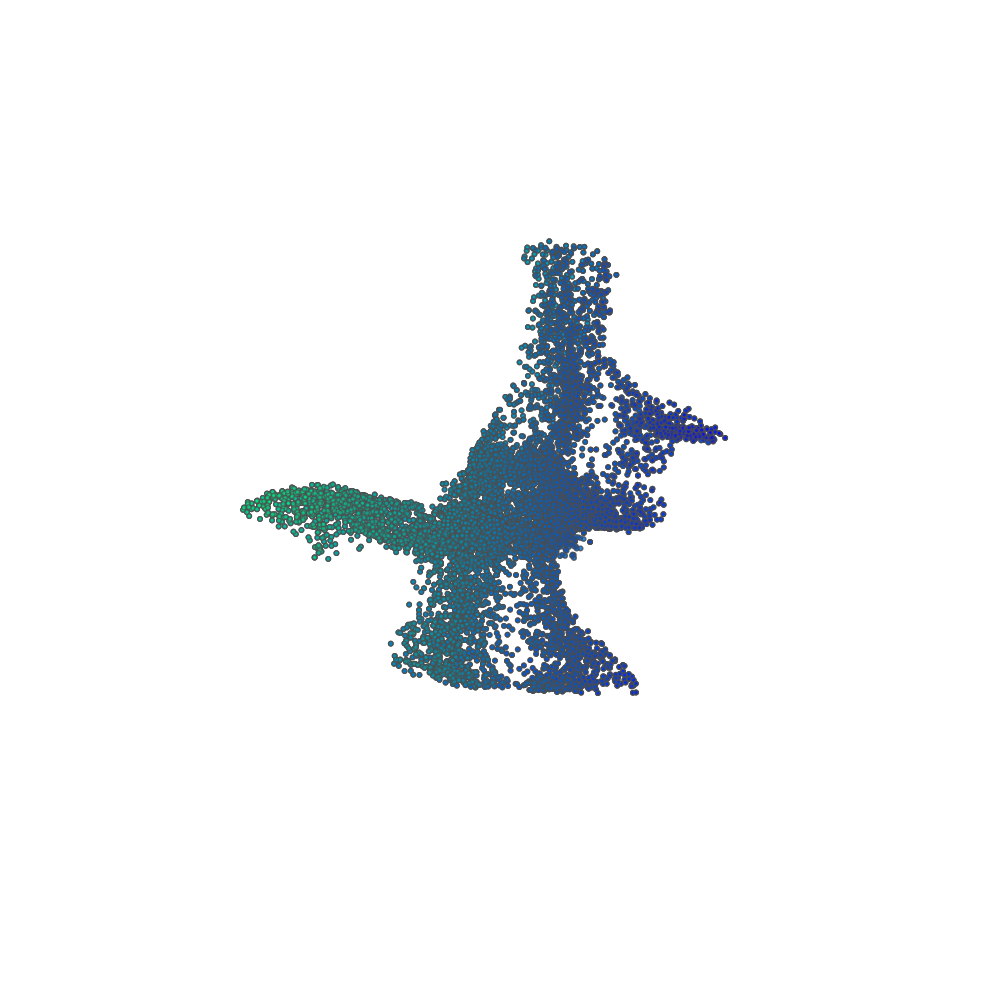}
              \caption{ADD with dual and triple object input.}
              \label{fig:multiple}
          \end{figure}
      \end{minipage}
  \end{minipage}

  In addition to using union in the transformation process in every iteration, 
  we can push this idea further by using amalgamated point clouds as input instead of a single object.
  The results of ADD with the union of two or three objects are shown in Figure~\ref{fig:multiple}. Compared with
  Figure~\ref{fig:single}, ADD with multiple inputs results in objects with more geometric details benefited by a more
  versatile input space. 
  \begin{wrapfigure}[9]{r}{0.40\textwidth}
  \vspace{1em}
  \includegraphics[width=0.49\linewidth]{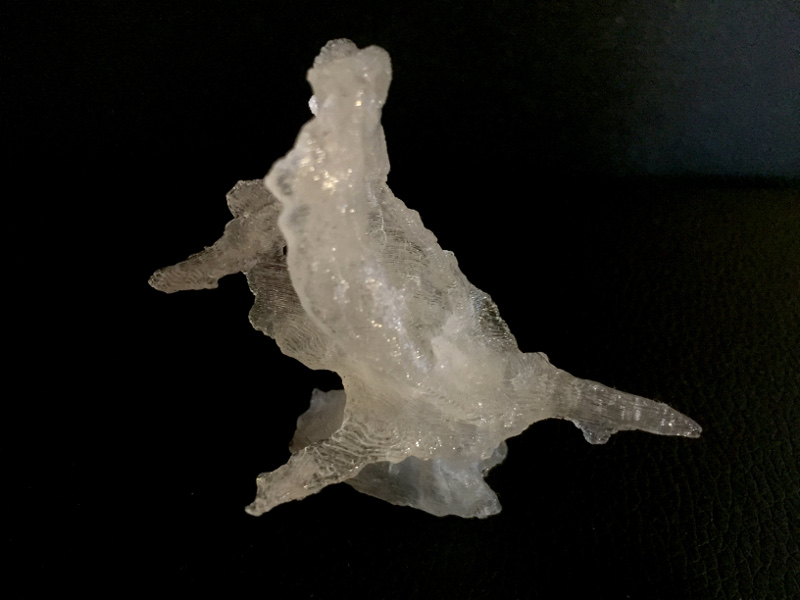}
  \hfill
  \includegraphics[width=0.49\linewidth]{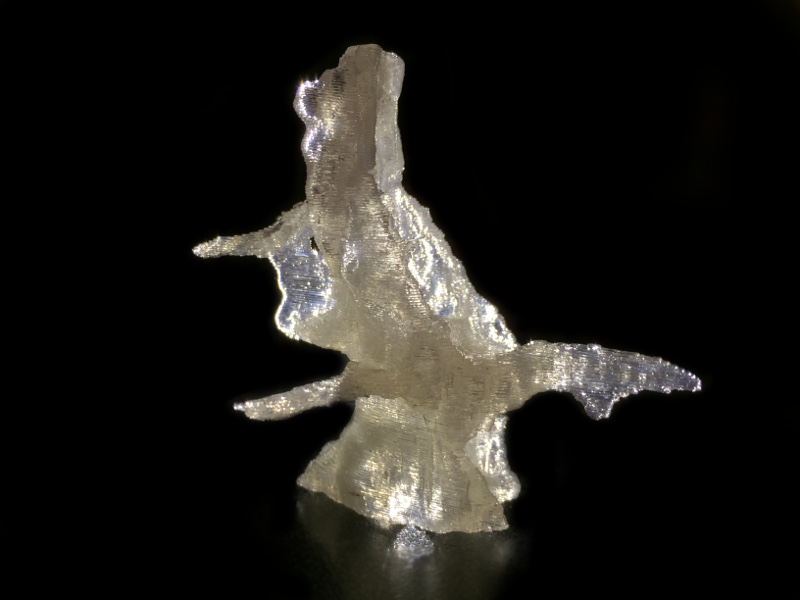}
  \caption{Created Sculpture from ADD.}
  \label{fig:print}
  \end{wrapfigure}

  \vspace{-10pt}
  \paragraph{From point clouds to realizable 3D sculptures}
  Our final goal is to create 3D sculptures in the real world. 
  We use standard software MeshLab and Meshmixer to reconstruct a mesh from point clouds
  created by ADD. We then use Lulzbot TAZ 3D printer with dual-extruder, transparent PLA, and dissolvable supporting material
  to create the sculpture of the reconstructed mesh. 
  The printed sculptures of the point cloud in Figure~\ref{fig:multiple} (RHS) are shown in Figure~\ref{fig:print}.

\Urlmuskip=0mu plus 1mu\relax
\bibliographystyle{plainnat}
\bibliography{main}

\begin{thebibliography}{21}
\providecommand{\natexlab}[1]{#1}
\providecommand{\url}[1]{\texttt{#1}}
\expandafter\ifx\csname urlstyle\endcsname\relax
  \providecommand{\doi}[1]{doi: #1}\else
  \providecommand{\doi}{doi: \begingroup \urlstyle{rm}\Url}\fi

\bibitem[dee()]{deepdream}
URL
  https://ai.googleblog.com/2015/06/inceptionism-going-deeper-into-neural.html.

\bibitem[one()]{one_human}
URL
  https://www.christies.com/features/A-collaboration-between-two-artists-one-human-one
  -a-machine-9332-1.aspx.

\bibitem[rob()]{robotart2018}
URL https://robotart.org/2018-winners/.

\bibitem[Achlioptas et~al.(2017)Achlioptas, Diamanti, Mitliagkas, and
  Guibas]{achlioptas2017learning}
Panos Achlioptas, Olga Diamanti, Ioannis Mitliagkas, and Leonidas Guibas.
\newblock Learning representations and generative models for 3d point clouds.
\newblock \emph{arXiv preprint arXiv:1707.02392}, 2017.

\bibitem[{Brown} et~al.(2018){Brown}, {Carlini}, {Zhang}, {Olsson},
  {Christiano}, and {Goodfellow}]{unrestricted_Advex_2018}
T.~B. {Brown}, N.~{Carlini}, C.~{Zhang}, C.~{Olsson}, P.~{Christiano}, and
  I.~{Goodfellow}.
\newblock Unrestricted adversarial examples.
\newblock \emph{arXiv preprint arXiv:1809.08352}, 2018.

\bibitem[Carter and Nielsen(2017)]{carter2017using}
Shan Carter and Michael Nielsen.
\newblock Using artificial intelligence to augment human intelligence.
\newblock \emph{Distill}, 2017.

\bibitem[Elgammal et~al.(2017)Elgammal, Liu, Elhoseiny, and
  Mazzone]{elgammal2017can}
Ahmed Elgammal, Bingchen Liu, Mohamed Elhoseiny, and Marian Mazzone.
\newblock Can: Creative adversarial networks, generating" art" by learning
  about styles and deviating from style norms.
\newblock \emph{arXiv preprint arXiv:1706.07068}, 2017.

\bibitem[Goodfellow et~al.(2014)Goodfellow, Pouget-Abadie, Mirza, Xu,
  Warde-Farley, Ozair, Courville, and Bengio]{goodfellow2014generative}
Ian Goodfellow, Jean Pouget-Abadie, Mehdi Mirza, Bing Xu, David Warde-Farley,
  Sherjil Ozair, Aaron Courville, and Yoshua Bengio.
\newblock Generative adversarial nets.
\newblock In \emph{NIPS}, 2014.

\bibitem[Groueix et~al.(2018)Groueix, Fisher, Kim, Russell, and
  Aubry]{groueix2018atlasnet}
Thibault Groueix, Matthew Fisher, Vladimir~G Kim, Bryan~C Russell, and Mathieu
  Aubry.
\newblock Atlasnet: A papier-mache approach to learning 3d surface generation.
\newblock \emph{arXiv preprint arXiv:1802.05384}, 2018.

\bibitem[Hoppe et~al.(1992)Hoppe, DeRose, Duchamp, McDonald, and
  Stuetzle]{hoppe1992surface}
Hugues Hoppe, Tony DeRose, Tom Duchamp, John McDonald, and Werner Stuetzle.
\newblock \emph{Surface reconstruction from unorganized points}.
\newblock 1992.

\bibitem[Kato et~al.(2018)Kato, Ushiku, and Harada]{kato2018neural}
Hiroharu Kato, Yoshitaka Ushiku, and Tatsuya Harada.
\newblock Neural 3d mesh renderer.
\newblock In \emph{Proceedings of the IEEE Conference on Computer Vision and
  Pattern Recognition}, pages 3907--3916, 2018.

\bibitem[Lehman et~al.(2016)Lehman, Risi, and Clune]{lehman2016creative}
Joel Lehman, Sebastian Risi, and Jeff Clune.
\newblock Creative generation of 3d objects with deep learning and innovation
  engines.
\newblock In \emph{Proceedings of the 7th International Conference on
  Computational Creativity}, 2016.

\bibitem[Li et~al.(2018)Li, Zaheer, Zhang, Poczos, and
  Salakhutdinov]{li2018point}
Chun-Liang Li, Manzil Zaheer, Yang Zhang, Barnabas Poczos, and Ruslan
  Salakhutdinov.
\newblock Point cloud gan.
\newblock \emph{arXiv preprint arXiv:1810.05795}, 2018.

\bibitem[Liu et~al.(2018)Liu, Tao, Li, Nowrouzezahrai, and
  Jacobson]{liu2018adversarial}
Hsueh-Ti~Derek Liu, Michael Tao, Chun-Liang Li, Derek Nowrouzezahrai, and Alec
  Jacobson.
\newblock Adversarial geometry and lighting using a differentiable renderer.
\newblock \emph{arXiv preprint arXiv:1808.02651}, 2018.

\bibitem[Qi et~al.(2017)Qi, Su, Mo, and Guibas]{qi2017pointnet}
Charles~R Qi, Hao Su, Kaichun Mo, and Leonidas~J Guibas.
\newblock Pointnet: Deep learning on point sets for 3d classification and
  segmentation.
\newblock \emph{CVPR}, 2017.

\bibitem[Szegedy et~al.(2013)Szegedy, Zaremba, Sutskever, Bruna, Erhan,
  Goodfellow, and Fergus]{szegedy2013intriguing}
Christian Szegedy, Wojciech Zaremba, Ilya Sutskever, Joan Bruna, Dumitru Erhan,
  Ian Goodfellow, and Rob Fergus.
\newblock Intriguing properties of neural networks.
\newblock \emph{arXiv preprint arXiv:1312.6199}, 2013.

\bibitem[Van Den~Oord et~al.(2016)Van Den~Oord, Dieleman, Zen, Simonyan,
  Vinyals, Graves, Kalchbrenner, Senior, and Kavukcuoglu]{van2016wavenet}
A{\"a}ron Van Den~Oord, Sander Dieleman, Heiga Zen, Karen Simonyan, Oriol
  Vinyals, Alex Graves, Nal Kalchbrenner, Andrew~W Senior, and Koray
  Kavukcuoglu.
\newblock Wavenet: A generative model for raw audio.
\newblock In \emph{SSW}, page 125, 2016.

\bibitem[Wu et~al.(2016)Wu, Zhang, Xue, Freeman, and Tenenbaum]{wu2016learning}
Jiajun Wu, Chengkai Zhang, Tianfan Xue, Bill Freeman, and Josh Tenenbaum.
\newblock Learning a probabilistic latent space of object shapes via 3d
  generative-adversarial modeling.
\newblock In \emph{NIPS}, 2016.

\bibitem[Wu et~al.(2015)Wu, Song, Khosla, Yu, Zhang, Tang, and Xiao]{wu20153d}
Zhirong Wu, Shuran Song, Aditya Khosla, Fisher Yu, Linguang Zhang, Xiaoou Tang,
  and Jianxiong Xiao.
\newblock 3d shapenets: A deep representation for volumetric shapes.
\newblock In \emph{CVPR}, 2015.

\bibitem[Yang et~al.(2018)Yang, Feng, Shen, and Tian]{yang2018foldingnet}
Yaoqing Yang, Chen Feng, Yiru Shen, and Dong Tian.
\newblock Foldingnet: Point cloud auto-encoder via deep grid deformation.
\newblock In \emph{CVPR}, volume~3, 2018.

\bibitem[Zaheer et~al.(2017)Zaheer, Kottur, Ravanbakhsh, Poczos, Salakhutdinov,
  and Smola]{zaheer2017deep}
Manzil Zaheer, Satwik Kottur, Siamak Ravanbakhsh, Barnabas Poczos, Ruslan~R
  Salakhutdinov, and Alexander~J Smola.
\newblock Deep sets.
\newblock In \emph{NIPS}, 2017.

\end{thebibliography}
\end{document}